\documentclass[11pt]{article}

\usepackage[preprint]{acl}

\usepackage{times}
\usepackage{latexsym}
\usepackage[T1]{fontenc}
\usepackage[utf8]{inputenc}
\usepackage{microtype}
\usepackage{graphicx}
\usepackage{amsmath}
\usepackage{amssymb}
\usepackage{booktabs}
\usepackage{multirow}
\usepackage{xspace}
\usepackage{algorithm}
\usepackage{algpseudocode}

\newcommand{\method}{SAE-FD\xspace}

\title{SAE-FD: Sparse Autoencoder Feature Distillation for \\ Continual Learning of Large Language Models}

\author{
  Mingxu Zhang\textsuperscript{1},
  Yuhan Li\textsuperscript{1},
  Lujundong Li\textsuperscript{1},
  Dazhong Shen\textsuperscript{2}\thanks{Corresponding authors.},
  Hui Xiong\textsuperscript{1}\thanks{Corresponding authors.},
  Ying Sun\textsuperscript{3}\footnotemark[1] \\
  \textsuperscript{1}The Hong Kong University of Science and Technology (Guangzhou) \\
  \textsuperscript{2}Nanjing University of Aeronautics and Astronautics \\
  \textsuperscript{3}The 63rd Research Institute, National University of Defense Technology, Nanjing\\
  \texttt{mzhang630@connect.hkust-gz.edu.cn}, 
  \texttt{shendazhong@nuaa.edu.cn}, \\
  \texttt{sunyinggilly@gmail.com}
}
\begin{document}
\maketitle

\begin{abstract}
Continual learning enables large language models to adapt to evolving tasks without retraining from scratch, yet catastrophic forgetting remains a central obstacle.
Among continual learning methods, regularization-based approaches are widely used to constrain model updates and reduce forgetting, operating in weight space, gradient space, or output space.
However, these dense representation spaces suffer from feature superposition, where multiple concepts are encoded in overlapping dimensions, making it difficult to selectively protect previously learned knowledge without impeding new-task learning.
To address this issue, we propose \method (Sparse Autoencoder Feature Distillation), which anchors model representations in the sparse feature space of a pre-trained Sparse Autoencoder, where dense activations are decomposed into a sparse overcomplete basis that reduces representational entanglement, enabling more targeted regularization with less interference to new-task learning.
Experiments on two continual learning benchmarks across three model architectures show that \method consistently outperforms existing regularization-based methods, achieving up to 52.70\% average accuracy with only $-$0.46\% backward transfer.
\end{abstract}

\section{Introduction}

Large language models (LLMs) have demonstrated strong capabilities across diverse domains, including molecular science \cite{zhang2025chematp, zhang2025atomdisc} and social science \cite{cui2026llm}. In practice, deploying LLMs for specific application scenarios often requires fine-tuning on specific data.
When models must be sequentially adapted to multiple tasks over time, catastrophic forgetting \cite{kirkpatrick2017ewc} becomes a central challenge: training on a new task degrades performance on previously learned tasks due to the overwriting and interference of internal representations.
To mitigate forgetting, existing continual learning (CL) methods regularize model updates in different representation spaces.
Weight-space methods such as EWC \cite{kirkpatrick2017ewc} penalize changes to important parameters; gradient-space methods such as O-LoRA \cite{wang2023olora} and GORP \cite{wang2025gorp} project updates to be orthogonal to previous task subspaces; output-space methods such as LwF \cite{li2017lwf} and SEEKR \cite{he2024seekr} distill model outputs or attention patterns.
A common limitation is that all three families operate in dense representation spaces where features representing different tasks are superimposed \cite{elhage2022superposition}, making it difficult to selectively protect old-task knowledge without constraining new-task learning.

Sparse Autoencoders (SAEs) \cite{cunningham2024sae, bricken2023monosemanticity}, developed for mechanistic interpretability, decompose dense activations into sparse, overcomplete feature vectors ($D \gg d$) that disentangle superimposed representations.
The high sparsity and the overcomplete basis provide a promising decoupled space for continual learning regularization, where the reduced entanglement allows more targeted protection of previously learned knowledge.
However, this application is non-trivial, as it requires training model-specific SAEs, designing loss functions that maintain gradient flow through the sparse encoding, and adapting regularization strength across different tasks and training stages.

To address these challenges, we introduce \method (Sparse Autoencoder Feature Distillation), a continual learning method that performs representation anchoring in SAE feature space.
\method operates in three stages.
First, we train a Gated SAE on diverse text activations from the base model to obtain a high-quality sparse feature decomposition.
Second, after fine-tuning on each task using LoRA, we capture per-token SAE feature activations on a small set of anchor samples, creating a compact snapshot of the model's learned representations.
Third, during subsequent task training, we apply a sparse-aware distillation loss that combines cosine direction preservation with active-feature magnitude matching.
The total training loss combines the task loss with the distillation loss weighted by a coefficient $\lambda$; since both losses vary in magnitude across tasks and training steps, a fixed $\lambda$ cannot maintain a consistent regularization effect.
We therefore introduce an adaptive mechanism that adjusts $\lambda$ to target a fixed contribution ratio of the distillation loss, providing strong protection at the start of each new task and relaxing as the model converges.
By operating in the sparse feature space, \method reduces the interference between regularization and new-task learning, achieving a better plasticity-stability tradeoff than dense-space alternatives.

In conclusion, our contributions are:
\begin{itemize}
\item We propose \method, the first continual learning method that performs regularization in the sparse, decoupled feature space of SAEs, enabling more targeted preservation of previously learned knowledge than baselines.
\item We design an adaptive $\lambda$ mechanism that dynamically adjusts regularization strength during training by targeting a fixed contribution ratio of the distillation loss to the total loss, providing stronger protection when forgetting risk is high and relaxing as the model stabilizes.
\item Experiments on two CL benchmarks across three model families show that \method achieves 52.70\% average accuracy with only $-$0.46\% backward transfer on TRACE, outperforming the strongest published baseline by +2.3\% AA with 34\% less forgetting.
\end{itemize}

\section{Related Work}

\paragraph{Continual Learning for LLMs.}
Parameter-efficient fine-tuning via LoRA \cite{hu2022lora} has become the dominant LLM adaptation strategy, enabling a family of LoRA-based CL methods.
O-LoRA \cite{wang2023olora} maintains orthogonal subspaces across tasks; GORP \cite{wang2025gorp} projects gradients into low-rank subspaces optimized for minimal interference; TreeLoRA \cite{qian2025treelora} organizes task-specific LoRA modules in a hierarchical structure with knowledge distillation; N-LoRA \cite{yang2025nlora} addresses parameter collision between tasks; InfLoRA \cite{liang2024inflora} constructs interference-free low-rank subspaces.
Prompt-based methods including L2P \cite{wang2022l2p}, DualPrompt \cite{wang2022dualprompt}, and Progressive Prompts \cite{razdaibiedina2023progressive} prepend learnable tokens to model inputs, but their effectiveness is limited on generation tasks.
Replay-based methods maintain a memory buffer of previous-task exemplars; SEEKR \cite{he2024seekr} combines selective attention distillation with replay, achieving strong results but requiring stored training data and full fine-tuning.

\paragraph{Representation Distillation for CL.}
A complementary line of work preserves previous-task knowledge through distillation at various levels of model internals.
At the output level, LwF \cite{li2017lwf} distills logits from a frozen teacher, and DER++ \cite{buzzega2020der} stores logits alongside replay exemplars.
At the attention level, SEEKR \cite{he2024seekr} identifies and distills the most task-relevant attention patterns, and DATA \cite{liao2025data} decomposes attention for task adaptation.
At the hidden state level, GeRe \cite{zhang2025gere} enforces activation consistency on anchor samples, demonstrating that feature-level distillation outperforms logit-level distillation, though their ablation shows that raw L1/L2 matching of dense activations is suboptimal.
\method distills in the sparse, decoupled feature space of an SAE, reducing representational entanglement before applying regularization.

\paragraph{Sparse Autoencoders.}
SAEs learn to decompose neural network activations into sparse, overcomplete features \cite{cunningham2024sae, bricken2023monosemanticity}, and recent variants such as Gated SAEs \cite{rajamanoharan2024gatedsae} achieve high-quality reconstruction with sparser features, scaling reliably to production-scale models \cite{templeton2024scaling, gao2024scaling}.
The superposition hypothesis \cite{elhage2022superposition}, which posits that networks represent more features than they have dimensions by exploiting sparsity, provides theoretical grounding for why SAE-based decomposition can disentangle overlapping representations.
Beyond interpretability, SAEs have been applied to controllable generation, where \citet{zhang2026slim} show that SAE features can serve as decoupled intervention handles for steering molecular properties in LLMs, demonstrating the practical utility of SAE-based disentanglement.
To our knowledge, no prior work has explored SAE feature space as a regularization target for continual learning.

\section{Preliminaries}
\label{sec:preliminaries}

\paragraph{Continual Learning.}
We consider a sequence of $T$ tasks $\{\mathcal{D}_1, \ldots, \mathcal{D}_T\}$, where each task $\mathcal{D}_t = \{(x_i^t, y_i^t)\}$ consists of input-output pairs.
The model is trained sequentially on each task.
A central challenge in this setting is catastrophic forgetting: as the model adapts to later tasks, performance on earlier tasks may degrade due to interference between task-specific representations.
To quantify this, we use two standard metrics.
Let $a_{i,t}$ denote the performance on task $t$ after training through task $i$:
\begin{align}
\text{AA}_T &= \frac{1}{T}\sum_{t=1}^{T} a_{T,t} \\
\text{BWT}_T &= \frac{1}{T-1}\sum_{t=1}^{T-1}(a_{T,t} - a_{t,t})
\end{align}
Average Accuracy (AA) measures the overall performance across all tasks after the final training stage; higher AA indicates stronger overall capability.
Backward Transfer (BWT) captures how much each task's performance degrades after subsequent training; a BWT of zero indicates no forgetting, and higher (less negative) BWT is better.
An ideal continual learner achieves both high AA (strong performance on all tasks) and BWT close to zero (no forgetting of previously learned tasks).

\paragraph{Sparse Autoencoder.}
Sparse Autoencoders (SAEs) \cite{cunningham2024sae} are dictionary learning models trained to decompose dense neural network activations into sparse linear combinations of learned feature directions.
Given an activation vector $\mathbf{h} \in \mathbb{R}^d$, an SAE maps it to a higher-dimensional feature space $\mathbb{R}^D$ ($D \gg d$), where the resulting feature vector is sparse: only a small fraction of features are active for any given input.
This overcomplete, sparse representation disentangles superimposed concepts in the original activation space, producing features that are more decoupled than the raw dimensions \cite{bricken2023monosemanticity}.

In this work, we adopt the Gated SAE variant \cite{rajamanoharan2024gatedsae}, which separates the encoding process into gating and magnitude pathways:
\begin{align}
\mathbf{g} &= \sigma(W_{\text{gate}} (\mathbf{h} - \mathbf{b}_{\text{dec}}) + \mathbf{b}_{\text{gate}}) \\
\mathbf{m} &= W_{\text{enc}} (\mathbf{h} - \mathbf{b}_{\text{dec}}) + \mathbf{b}_{\text{enc}} \\
\mathbf{f}_{\text{pre}} &= \mathbf{g} \odot \mathbf{m} \label{eq:pre_relu} \\
\mathbf{f} &= \text{ReLU}(\mathbf{f}_{\text{pre}}) \label{eq:post_relu}
\end{align}
where $\sigma$ is the sigmoid function, $W_{\text{gate}}, W_{\text{enc}} \in \mathbb{R}^{D \times d}$, $\odot$ denotes element-wise multiplication, and $\mathbf{b}_{\text{dec}} \in \mathbb{R}^d$ is the decoder bias used for input centering.
The sigmoid gate $\mathbf{g}$ determines which features are active, while the magnitude pathway $\mathbf{m}$ computes the feature strengths.
The post-ReLU features $\mathbf{f}$ are truly sparse, while the pre-ReLU features $\mathbf{f}_{\text{pre}}$ retain gradient flow through the sigmoid gate.
Reconstruction is performed by $\hat{\mathbf{h}} = W_{\text{dec}} \mathbf{f} + \mathbf{b}_{\text{dec}}$, where $W_{\text{dec}} \in \mathbb{R}^{D \times d}$ has unit-norm rows.

\begin{figure*}[t]
\centering
\includegraphics[width=\textwidth]{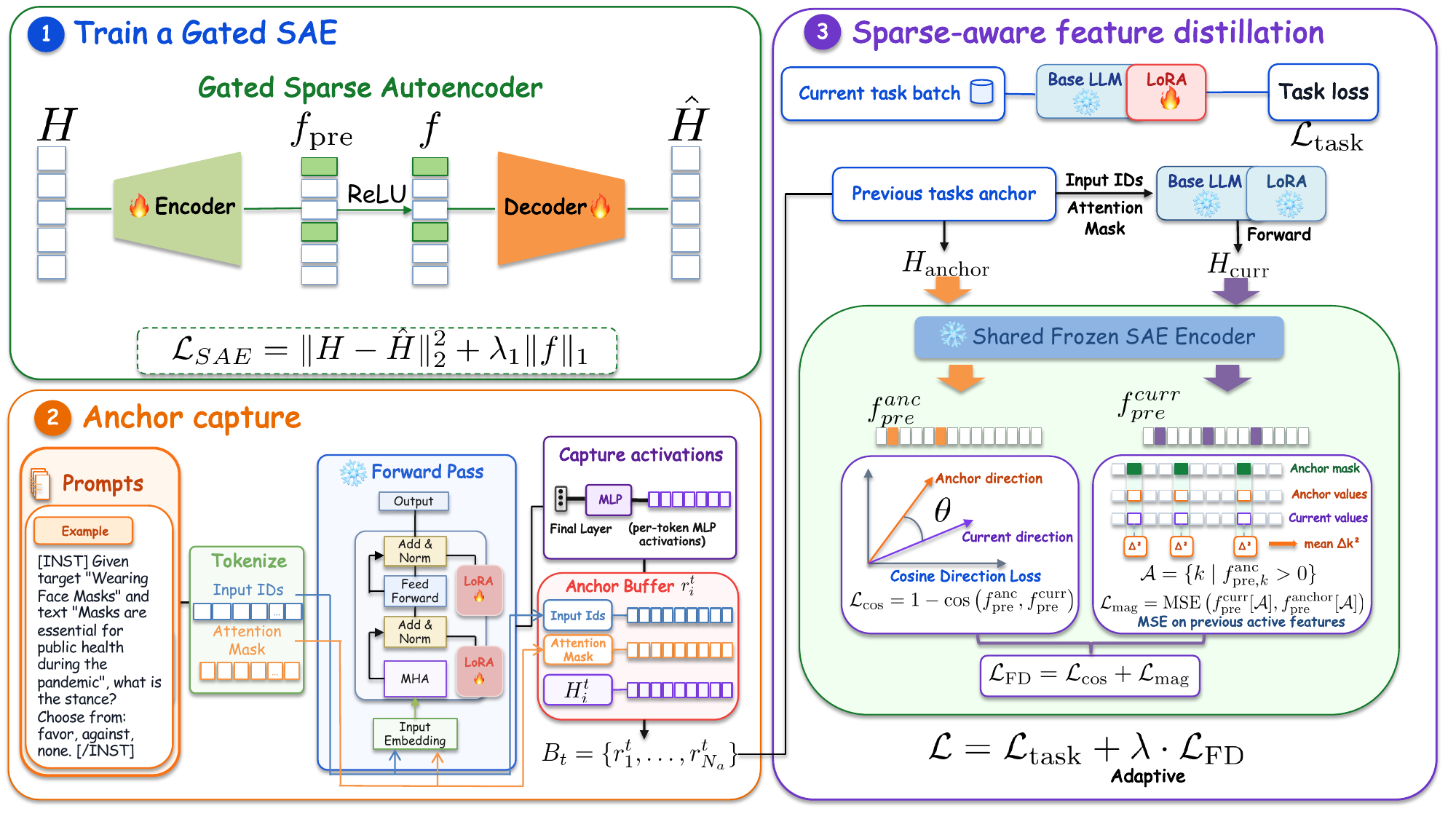}
\caption{Overview of \method. \textbf{Stage 1}: A Gated SAE is trained to decompose dense activations into sparse features. \textbf{Stage 2}: After each task's training, per-token MLP activations on anchor samples are captured and stored. \textbf{Stage 3}: Stored anchor activations and recomputed current activations are encoded through the frozen SAE, and the sparse-aware distillation loss combines cosine direction preservation with active-feature magnitude matching, weighted by an adaptive $\lambda$.}
\label{fig:overview}
\end{figure*}

\section{Method}
\label{sec:method}

\method operates as a three-stage pipeline within a continual learning loop: (1)~training a Gated SAE to decompose model activations into sparse features, (2)~training on each task with anchor capture, and (3)~applying sparse-aware feature distillation during subsequent tasks.
Algorithm~\ref{alg:saefd} provides the complete procedure.

\subsection{Stage 1: SAE Training}
\label{sec:sae_training}

As shown in Figure~\ref{fig:overview} (Stage 1), before the continual learning sequence begins, we train a Gated SAE to decompose the base model's internal representations into sparse features.
This SAE is trained once per base model and remains frozen throughout the continual learning process.

\paragraph{Activation collection.}
We collect last-layer MLP activations from the base model on diverse text sources, including classification datasets (AG News, Amazon, Yelp, DBpedia, Yahoo), question answering (BoolQ, ARC), and knowledge benchmarks (MMLU), with up to 2{,}000 samples per source.
For each input, we extract the last non-padding token's MLP output from the final transformer layer, yielding a dataset of $d$-dimensional activation vectors that captures the model's representation diversity.

\paragraph{Training objective.}
The SAE is trained with a reconstruction loss and an L1 sparsity penalty:
\begin{equation}
\mathcal{L}_{\text{SAE}} = \| \hat{\mathbf{h}} - \mathbf{h} \|_2^2 + \lambda_{\ell_1} \| \mathbf{f} \|_1
\label{eq:sae_loss}
\end{equation}
where $\hat{\mathbf{h}} = W_{\text{dec}} \mathbf{f} + \mathbf{b}_{\text{dec}}$ is the reconstruction and $\mathbf{f}$ is the post-ReLU feature vector (Eq.~\ref{eq:post_relu}).
The reconstruction term ensures the SAE faithfully represents the original activations, while the L1 term encourages sparse feature usage, activating only the features relevant to each input.
We use $d{=}4096$ and $D{=}32768$ (8$\times$ expansion), with $\lambda_{\ell_1} = 10^{-3}$.
Each trained SAE achieves $>$99\% variance explained on held-out activations.

\subsection{Stage 2: Task Training and Anchor Capture}
\label{sec:anchor_capture}

As shown in Figure~\ref{fig:overview} (Stage 2), this stage serves two purposes: fine-tuning $\mathcal{M}$ on each task, and capturing a representation snapshot in SAE feature space that anchors the model's learned knowledge for subsequent distillation. For each task $t$ in the continual learning sequence, we fine-tune $\mathcal{M}$ on $\mathcal{D}_t$ using LoRA \cite{hu2022lora} with the standard language modeling objective:
\begin{equation}
\mathcal{L}_{\text{task}} = -\sum_{j} \log P_{\mathcal{M}}(y_j \mid x_{<j})
\end{equation}
where $y_j$ denotes the target tokens.
When $t > 1$, the training loss additionally includes the feature distillation term described in \S\ref{sec:fd_loss}.

\paragraph{Anchor capture.}
To provide a reference for subsequent distillation in SAE feature space, we capture a compact snapshot of the model's representations after training on each task.
Specifically, we select $N_a$ anchor samples from $\mathcal{D}_t$ and perform a forward pass through $\mathcal{M}$.
For each anchor input $x_i = (x_1, \ldots, x_L)$ with attention mask $\mathbf{m}_i$, we extract the per-token MLP output from the last transformer layer, yielding $\mathbf{H}_i \in \mathbb{R}^{L \times d}$.
The anchor tuple $(x_i, \mathbf{m}_i, \mathbf{H}_i)$ is stored in float16 for memory efficiency (${\sim}$0.8~GB per task).
During future training, both $\mathbf{H}_i$ and the current model's activations on $x_i$ are encoded through the frozen SAE $\mathcal{S}$ to compute the distillation loss, enabling the model to be regularized in the decoupled feature space rather than in the raw activation space.

\paragraph{Anchor accumulation.}
The anchor buffer accumulates across tasks: when training on task $t{+}1$, the buffer contains anchors from all tasks $1, \ldots, t$.
At each training step, one anchor batch is sampled uniformly at random from the buffer, amortizing the distillation cost across all previous tasks.

\subsection{Stage 3: Sparse-Aware Feature Distillation}
\label{sec:fd_loss}

As shown in Figure~\ref{fig:overview} (Stage 3), as the model trains on task $t{+}1$, its internal representations inevitably drift from those learned for previous tasks.
The goal of this stage is to constrain this drift in SAE feature space, where the decoupled structure allows us to measure and penalize representational changes more precisely than in the raw activation space.
Concretely, we encode both the stored anchor activations $\mathbf{H}_a$ and the model's current activations on the same input through the frozen SAE $\mathcal{S}$, and compare the resulting feature vectors.
Both encodings use the pre-ReLU features (Eq.~\ref{eq:pre_relu}) to ensure gradient flow through the sigmoid gate; post-ReLU features would zero out gradients for inactive features.

Representation drift in SAE space manifests in two complementary ways: the overall feature direction may rotate, and individual feature magnitudes may shift.
We address each with a dedicated loss term.

\paragraph{Cosine direction loss.}
To prevent the global feature direction from drifting, we penalize the angular deviation between the current feature vector $\mathbf{f}_{\text{pre}}^{\text{curr}}$ produced by encoding the model's current activations and the anchor feature vector $\mathbf{f}_{\text{pre}}^{\text{anc}}$ produced by encoding the stored activations, both through the frozen SAE:
\begin{equation}
\mathcal{L}_{\text{cos}} = 1 - \frac{\mathbf{f}_{\text{pre}}^{\text{curr}} \cdot \mathbf{f}_{\text{pre}}^{\text{anc}}}{\|\mathbf{f}_{\text{pre}}^{\text{curr}}\| \cdot \|\mathbf{f}_{\text{pre}}^{\text{anc}}\|}
\end{equation}
This loss is computed only on tokens whose anchor feature norm exceeds a threshold (10\% of the batch median norm), avoiding numerical instability from near-zero vectors.

\paragraph{Active magnitude loss.}
While the cosine loss preserves the global direction, it is scale-invariant and cannot prevent individual feature magnitudes from drifting.
We therefore add a magnitude loss that applies MSE selectively on features that were active (post-ReLU $> 0$) in the anchor, directly constraining the features that contributed to the model's previous-task behavior while leaving inactive features free to be recruited for new tasks:
\begin{equation}
\mathcal{L}_{\text{mag}} = \frac{1}{|\mathcal{A}|}\sum_{k \in \mathcal{A}} (f_{\text{pre},k}^{\text{curr}} - f_{\text{pre},k}^{\text{anc}})^2
\end{equation}
where $\mathcal{A} = \{k : f_k^{\text{anc}} > 0\}$ is the sparse set of previously active features.
By focusing regularization only on these active features, this loss directly leverages the sparse, decoupled structure of SAE representations.
Inactive features are left entirely unconstrained, preserving the model's capacity to recruit new features for upcoming tasks.

\paragraph{Combined loss.}
The feature distillation loss $\mathcal{L}_{\text{FD}} = \mathcal{L}_{\text{cos}} + \mathcal{L}_{\text{mag}}$ is averaged over non-padding tokens in the sampled anchor batch.
The cosine term preserves the global direction of the feature representation while the magnitude term ensures individual active features retain their strengths, jointly maintaining both the structure and the scale of previously learned representations.

\subsection{Adaptive $\lambda$ Mechanism}
\label{sec:adaptive_lambda}

The total training objective for task $t{+}1$ combines the task loss with the feature distillation loss:
\begin{equation}
\mathcal{L} = \mathcal{L}_{\text{task}} + \lambda \cdot \mathcal{L}_{\text{FD}}
\end{equation}
A key challenge is setting the coefficient $\lambda$.
A fixed $\lambda$ is problematic because the magnitudes of $\mathcal{L}_{\text{task}}$ and $\mathcal{L}_{\text{FD}}$ vary substantially across tasks and training steps: $\mathcal{L}_{\text{task}}$ is typically large at the start of a new task and decreases as the model converges, while $\mathcal{L}_{\text{FD}}$ depends on how much the current representations have drifted from the anchors.
A fixed $\lambda$ therefore cannot maintain a consistent balance between learning and regularization throughout training.

We address this with an adaptive mechanism that maintains a target contribution ratio $\rho$ of the distillation loss to the total loss.
Given the current losses, the ideal $\lambda$ satisfying $\rho = \frac{\lambda \mathcal{L}_{\text{FD}}}{\lambda \mathcal{L}_{\text{FD}} + \mathcal{L}_{\text{task}}}$ is:
\begin{equation}
\lambda^* = \frac{\rho}{1 - \rho} \cdot \frac{\mathcal{L}_{\text{task}}}{\mathcal{L}_{\text{FD}}}
\end{equation}
To avoid abrupt changes, $\lambda$ is updated via exponential moving average with clipping:
\begin{equation}
\lambda_{t+1} = (1 - \alpha) \lambda_t + \alpha \cdot \text{clip}(\lambda^*, \lambda_{\min}, \lambda_{\max})
\end{equation}
This design produces an intuitive behavior: since $\mathcal{L}_{\text{task}}$ is high when a new task begins, $\lambda$ automatically starts large, providing strong anti-forgetting protection precisely when the model is most vulnerable to catastrophic forgetting.
As the model converges on the new task, $\mathcal{L}_{\text{task}}$ decreases and $\lambda$ relaxes accordingly, allowing the model to fine-tune without excessive constraint.
The mechanism adapts to each task's learning dynamics without manual tuning.
Hyperparameter values ($\rho$, $\alpha$, $\lambda_{\min}$, $\lambda_{\max}$) are reported in Appendix~\ref{sec:appendix_settings}.

\begin{algorithm}[t]
\caption{\method Continual Learning Pipeline}
\label{alg:saefd}
\begin{algorithmic}[1]
\Require Task sequence $\{\mathcal{D}_1, \ldots, \mathcal{D}_T\}$, model $\mathcal{M}$, trained SAE $\mathcal{S}$, anchor count $N_a$
\State $\mathcal{A} \gets \emptyset$ \Comment{Anchor buffer}
\For{$t = 1, \ldots, T$}
    \For{each training step on $\mathcal{D}_t$}
        \State $\mathcal{L}_{\text{task}} \gets \text{CrossEntropy}(\mathcal{M}(x), y)$
        \If{$\mathcal{A} \neq \emptyset$}
            \State Sample $(x_a, \mathbf{m}_a, \mathbf{H}_a)$ from $\mathcal{A}$
            \State $\mathbf{H}_{\text{curr}} \gets \text{MLP}_{\text{last}}(\mathcal{M}, x_a)$
            \State $\mathbf{f}_{\text{pre}}^{\text{curr}} \gets \mathcal{S}.\text{encode\_pre\_relu}(\mathbf{H}_{\text{curr}})$
            \State $\mathbf{f}_{\text{pre}}^{\text{anc}} \gets \mathcal{S}.\text{encode\_pre\_relu}(\mathbf{H}_a)$
            \State $\mathcal{L}_{\text{FD}} \gets \mathcal{L}_{\text{cos}} + \mathcal{L}_{\text{mag}}$ \Comment{Eq.~7--8}
            \State Update $\lambda$ via EMA \Comment{Eq.~10}
            \State $\mathcal{L} \gets \mathcal{L}_{\text{task}} + \lambda \cdot \mathcal{L}_{\text{FD}}$
        \Else
            \State $\mathcal{L} \gets \mathcal{L}_{\text{task}}$
        \EndIf
        \State Update LoRA parameters via $\nabla \mathcal{L}$
    \EndFor
    \State \textbf{Anchor capture:}
    \State Select $N_a$ samples from $\mathcal{D}_t$
    \For{each anchor sample $x_i$}
        \State $\mathbf{H}_i \gets \text{MLP}_{\text{last}}(\mathcal{M}, x_i)$ \Comment{Per-token}
        \State Store $(x_i, \mathbf{m}_i, \mathbf{H}_i^{\text{fp16}})$ in $\mathcal{A}$
    \EndFor
\EndFor
\end{algorithmic}
\end{algorithm}

\section{Experiments}
\label{sec:experiments}

We design experiments to answer the following research questions:
\textbf{RQ1}: Does \method effectively reduce catastrophic forgetting compared to existing CL methods across different benchmarks?
\textbf{RQ2}: Does the approach generalize across model architectures?
\textbf{RQ3}: How do the individual components (loss design, adaptive $\lambda$, SAE decomposition) contribute to performance?
\textbf{RQ4}: How sensitive is \method to its key hyperparameters?

\subsection{Setup}

\paragraph{Benchmarks.}
We primarily evaluate on \textbf{TRACE} \cite{wang2023trace}, a challenging CL benchmark that requires sequential adaptation across 8 heterogeneous tasks---from stance detection and financial sentiment to code completion and German text simplification---with 5{,}000 training and 500 test samples per task.
The diversity of task types (classification, generation, reasoning, multilingual) makes TRACE particularly suited for evaluating whether a CL method can preserve knowledge across dissimilar domains.
We additionally report results on a \textbf{Standard CL} 4-task text classification benchmark \cite{wang2025gorp, wang2023olora} in Appendix~\ref{sec:appendix_standard_cl}.
Detailed task descriptions, example prompts, and per-task evaluation metrics are provided in Appendix~\ref{sec:appendix_benchmark} and \ref{sec:appendix_metrics}.

\paragraph{Models and baselines.}
We evaluate across LLaMA-2-7B-Chat \cite{touvron2023llama2}, Vicuna-7B-v1.5 \cite{chiang2023vicuna}, and Mistral-7B-Instruct-v0.3 \cite{jiang2023mistral}, each with a separately trained Gated SAE.
Baselines include SeqLoRA, EWC \cite{kirkpatrick2017ewc}, O-LoRA \cite{wang2023olora}, GORP \cite{wang2025gorp}, TreeLoRA \cite{qian2025treelora}, L2P \cite{wang2022l2p}, DualPrompt \cite{wang2022dualprompt}, GEM \cite{lopezpaz2017gem}, OGD \cite{farajtabar2020ogd}, HiDe-PET \cite{wang2024hidepet}, LwF \cite{li2017lwf}, SEEKR \cite{he2024seekr}, PP \cite{razdaibiedina2023progressive}, LFPT5 \cite{qin2022lfpt5}, and DER++ \cite{buzzega2020der}.
We use published results where available; our reproduced baselines follow the official TRACE configuration.
Full training details and hyperparameters are in Appendix~\ref{sec:appendix_settings}.

\subsection{Main Results}

\paragraph{SAE-FD achieves the best plasticity-stability tradeoff among all compared methods.}
Table~\ref{tab:trace_main} shows that \method reaches 52.70\% AA with only $-$0.46\% BWT on LLaMA-2-7B-Chat.
Compared to GORP, the strongest published regularization-based method, \method improves accuracy by +2.3\% and reduces forgetting by 34\% ($-$0.46 vs $-$0.70 BWT).
The gap is more striking against other LoRA-based methods: EWC, O-LoRA, and TreeLoRA all show BWT worse than $-$3\%, indicating that weight-space and gradient-space regularization struggle to prevent forgetting on this diverse benchmark.
This validates our core hypothesis that sparse feature space provides a more effective regularization target than dense spaces where task features are superimposed.
A per-task BWT breakdown (Table~\ref{tab:bwt_breakdown}) shows that five of seven tasks exhibit $|\text{BWT}| < 2\%$, and the full accuracy matrix (Table~\ref{tab:accuracy_matrix}) confirms stable performance across the entire 8-task sequence.

\begin{table}[t]
\centering
\small
\setlength{\tabcolsep}{4pt}
\begin{tabular}{lcc}
\toprule
\textbf{Method} & \textbf{AA}$\uparrow$ & \textbf{BWT}$\uparrow$ \\
\midrule
\multicolumn{3}{l}{\textit{Prompt-based}} \\
PP \cite{razdaibiedina2023progressive} & 29.41 & $-$5.79 \\
L2P \cite{wang2022l2p} & 36.23 & $-$8.25 \\
DualPrompt \cite{wang2022dualprompt} & 37.69 & $-$8.03 \\
LFPT5 \cite{qin2022lfpt5} & 38.67 & $-$11.43 \\
\midrule
\multicolumn{3}{l}{\textit{LoRA / Regularization / Gradient}} \\
SeqLoRA & 34.30 & $-$18.50 \\
GEM \cite{lopezpaz2017gem} & 40.08 & $-$6.77 \\
HiDe-PET \cite{wang2024hidepet} & 41.60 & $-$7.12 \\
LwF \cite{li2017lwf} & 41.86 & $-$6.50 \\
OGD \cite{farajtabar2020ogd} & 42.09 & $-$8.06 \\
EWC \cite{kirkpatrick2017ewc} & 42.36 & $-$5.97 \\
O-LoRA \cite{wang2023olora} & 42.78 & $-$7.16 \\
TreeLoRA \cite{qian2025treelora} & 43.52 & $-$3.46 \\
GORP \cite{wang2025gorp} & 50.40 & $-$0.70 \\
\midrule
\method (ours) & \textbf{52.70} & \textbf{$-$0.46} \\
\bottomrule
\end{tabular}
\caption{TRACE benchmark results on LLaMA-2-7B-Chat. AA and BWT are reported in \%. \method achieves the best AA and BWT among all compared methods.}
\label{tab:trace_main}
\end{table}

\paragraph{SAE-FD generalizes to standard classification CL.}
To evaluate beyond the heterogeneous TRACE benchmark, we test on a 4-task text classification benchmark across 3 random task orders (Table~\ref{tab:standard_cl}).
\method achieves 88.25\% AA with $-$0.89\% BWT, reducing forgetting by 5.3$\times$ compared to SeqLoRA ($-$4.71\%), and shows low variance across orderings (AA ranges from 87.90\% to 88.55\%), suggesting robustness to task ordering.

\subsection{Multi-Model Generalization}

\paragraph{SAE-FD consistently outperforms published baselines across architectures.}
Table~\ref{tab:multi_model} compares \method against the strongest published baselines for each model.
On LLaMA-2, \method (52.70\%) outperforms GORP (50.40\%), the strongest published regularization-based method, by +2.3\% AA with 34\% less forgetting.
On Vicuna, \method (53.50\%) surpasses O-LoRA (43.42\%) by +10.1\% AA and achieves the best BWT among all methods ($-$0.69\%), including full fine-tuning baselines.
On Mistral, \method (57.80\%) surpasses TreeLoRA (54.77\%), the strongest published result, by +3.0\% AA with 2.3$\times$ less forgetting.
These consistent improvements across three architectures, each using a separately trained SAE with no method-level changes, demonstrate that the sparse feature decomposition principle generalizes beyond a single model family.
Notably, the gains are largest on Vicuna (+10.1\% AA over O-LoRA), where dense-space baselines degrade severely, suggesting that SAE-based regularization is particularly effective when the base model's representations are more susceptible to forgetting.

\begin{table}[t]
\centering
\small
\setlength{\tabcolsep}{3.5pt}
\begin{tabular}{llcc}
\toprule
\textbf{Model} & \textbf{Method} & \textbf{AA}$\uparrow$ & \textbf{BWT}$\uparrow$ \\
\midrule
\multirow{5}{*}{\shortstack[l]{LLaMA-2\\-7B}}
& EWC \cite{kirkpatrick2017ewc} & 42.36 & $-$5.97 \\
& O-LoRA \cite{wang2023olora} & 42.78 & $-$7.16 \\
& TreeLoRA \cite{qian2025treelora} & 43.52 & $-$3.46 \\
& GORP \cite{wang2025gorp} & 50.40 & $-$0.70 \\
& \method (ours) & \textbf{52.70} & \textbf{$-$0.46} \\
\midrule
\multirow{4}{*}{\shortstack[l]{Vicuna\\-7B}}
& LwF \cite{li2017lwf} & 41.19 & $-$5.54 \\
& EWC \cite{kirkpatrick2017ewc} & 41.88 & $-$15.57 \\
& O-LoRA \cite{wang2023olora} & 43.42 & $-$6.27 \\
& \method (ours) & \textbf{53.50} & $\mathbf{-0.69}$ \\
\midrule
\multirow{4}{*}{\shortstack[l]{Mistral\\-7B}}
& O-LoRA \cite{wang2023olora} & 52.02 & $-$8.13 \\
& EWC \cite{kirkpatrick2017ewc} & 52.45 & $-$5.98 \\
& TreeLoRA \cite{qian2025treelora} & 54.77 & $-$3.77 \\
& \method (ours) & \textbf{57.80} & $\mathbf{-1.63}$ \\
\bottomrule
\end{tabular}
\caption{Multi-model TRACE results compared with published baselines. Each model uses a separately trained SAE. AA and BWT are reported in \%. \method consistently achieves the best performance across all three architectures. Full baseline tables are provided in Appendix~\ref{sec:appendix_baselines}.}
\label{tab:multi_model}
\end{table}

\subsection{Ablation Studies}

To understand the contribution of each design choice, we conduct ablation studies on TRACE (Table~\ref{tab:ablations}).

\paragraph{Loss components reveal complementary roles.}
To examine whether both loss terms are necessary, we train \method with each component alone.
As shown in Table~\ref{tab:ablations}, cosine-only loss achieves 51.29\% AA with $-$0.51\% BWT---preserving feature directions but lacking magnitude control leads to lower accuracy.
Magnitude-only loss improves AA to 52.57\% but BWT degrades to $-$0.84\%, as matching magnitudes without direction constraints allows the overall representation to rotate.
The combined loss achieves the best on both metrics (52.70\% AA, $-$0.46\% BWT), confirming that direction and magnitude preservation address complementary aspects of representation drift.

\paragraph{Adaptive $\lambda$ is critical for strong anti-forgetting.}
To evaluate the adaptive mechanism, we compare it against a fixed $\lambda{=}1.0$.
As shown in Table~\ref{tab:ablations}, the fixed setting yields $-$2.38\% BWT, comparable to other regularization methods, while the adaptive mechanism achieves $-$0.46\%, a 5.2$\times$ improvement.
The dynamics (Appendix~\ref{sec:appendix_lambda}) reveal intuitive behavior: $\lambda$ starts at the maximum when beginning a new task and decays at task-dependent rates, e.g., short tasks like FOMC (3 epochs) decay rapidly ($10.0 \to 0.20$) while longer tasks like MeetingBank (7 epochs) maintain elevated $\lambda$ ($10.0 \to 4.80$).

\paragraph{SAE decomposition outperforms raw activation matching.}
To isolate the benefit of operating in SAE feature space, we compare \method against raw MSE distillation on the original 4096-dimensional MLP activations.
As shown in Table~\ref{tab:ablations}, SAE-FD consistently outperforms raw MSE: on Vicuna by +2.19\% AA ($-$0.69 vs $-$0.96 BWT), and on Mistral by +1.86\% BWT ($-$1.47 vs $-$3.33), confirming that the decoupled SAE space provides a more effective regularization target than the entangled dense space.
We additionally verify that the frozen SAE maintains faithful reconstruction quality ($>$90\% variance explained) throughout the 8-task sequence (Figure~\ref{fig:sae_quality}), and that the computational overhead is approximately 1.85$\times$ SeqLoRA with constant cost per step regardless of buffer size (Table~\ref{tab:efficiency}).

\begin{table}[t]
    \centering
    \small
    \setlength{\tabcolsep}{3.5pt}
    \begin{tabular}{llcc}
    \toprule
    \textbf{Ablation} & \textbf{Setting} & \textbf{AA}$\uparrow$ & \textbf{BWT}$\uparrow$ \\
    \midrule
    \multirow{3}{*}{\shortstack[l]{Loss\\(LLaMA-2)}} & Cosine only & 51.29 & -0.51 \\
    & Magnitude only & 52.57 & $-$0.84 \\
    & \textbf{Both} & \textbf{52.70} & $\mathbf{-0.46}$ \\
    \midrule
    \multirow{2}{*}{\shortstack[l]{$\lambda$\\(LLaMA-2)}} & Fixed $\lambda{=}1.0$ & 51.41 & $-$2.38 \\
    & \textbf{Adaptive} & \textbf{52.70} & $\mathbf{-0.46}$ \\
    \midrule
    \multirow{2}{*}{\shortstack[l]{SAE vs.\ MSE\\(Vicuna)}} & Raw MSE & 51.31 & $-$0.96 \\
    & \textbf{SAE-FD} & \textbf{53.50} & $\mathbf{-0.69}$ \\
    \midrule
    \multirow{2}{*}{\shortstack[l]{SAE vs.\ MSE\\(Mistral)}} & Raw MSE & 56.80 & $-$3.33 \\
    & \textbf{SAE-FD} & \textbf{57.13} & $\mathbf{-1.47}$ \\
    \bottomrule
    \end{tabular}
    \caption{Ablation studies on TRACE. AA and BWT are reported in \%. Each component contributes to the overall performance, and SAE-based distillation consistently outperforms raw MSE matching.}
    \label{tab:ablations}
\end{table}

\subsection{Hyperparameter Sensitivity}

We evaluate the sensitivity of \method to its two key hyperparameters: the number of anchor samples per task ($N_a$) and the adaptive $\lambda$ target ratio ($\rho$).
All experiments are conducted on TRACE with LLaMA-2-7B-Chat.

\paragraph{Anchor count $N_a$.}
Table~\ref{tab:sens_anchor} shows results for $N_a \in \{50, 100, 200, 400\}$.
Increasing the anchor count consistently improves BWT (from $-$1.45\% at $N_a{=}50$ to $-$0.22\% at $N_a{=}400$), as more anchor samples provide a more representative snapshot of previous-task representations.
AA remains stable across all settings (51.65--53.45\%), and all configurations substantially outperform SeqLoRA (AA=38.21\%).
We use $N_a{=}200$ as the default, balancing storage cost (${\sim}$0.8~GB/task) and anti-forgetting effectiveness.

\begin{table}[t]
\centering
\small
\begin{tabular}{lcc}
\toprule
\textbf{Anchor/task} & \textbf{AA (\%)}$\uparrow$ & \textbf{BWT (\%)}$\uparrow$ \\
\midrule
50 & 52.43 & $-$1.45 \\
100 & 51.65 & $-$0.66 \\
\textbf{200 (default)} & \textbf{52.70} & $\mathbf{-0.46}$ \\
400 & 53.45 & $-$0.22 \\
\bottomrule
\end{tabular}
\caption{Sensitivity to anchor count $N_a$ on TRACE (LLaMA-2-7B-Chat). Increasing $N_a$ consistently improves BWT while AA remains stable, indicating that a more representative anchor snapshot strengthens anti-forgetting without hurting new-task learning.}
\label{tab:sens_anchor}
\end{table}

\paragraph{Target ratio $\rho$.}
Table~\ref{tab:sens_rho} shows results for $\rho \in \{0.05, 0.10, 0.15, 0.25, 0.35\}$.
The method is robust across a wide range of $\rho$ values: AA ranges from 51.50\% to 54.35\% and BWT from $-$0.94\% to +0.63\%, all substantially better than SeqLoRA.
We use $\rho{=}0.15$ as the default, which provides a good tradeoff without overly constraining plasticity.

\begin{table}[t]
\centering
\small
\begin{tabular}{lcc}
\toprule
\textbf{Target ratio $\rho$} & \textbf{AA (\%)}$\uparrow$ & \textbf{BWT (\%)}$\uparrow$ \\
\midrule
0.05 & 51.50 & $-$0.94 \\
0.10 & 52.09 & +0.06 \\
\textbf{0.15 (default)} & \textbf{52.70} & $\mathbf{-0.46}$ \\
0.25 & 53.07 & $-$0.51 \\
0.35 & 54.35 & +0.63 \\
\bottomrule
\end{tabular}
\caption{Sensitivity to adaptive $\lambda$ target ratio $\rho$ on TRACE (LLaMA-2-7B-Chat). The method is robust across a wide range; all settings outperform SeqLoRA (AA=38.21\%).}
\label{tab:sens_rho}
\end{table}

Standard CL benchmark results and the full table with published baselines are provided in Appendix~\ref{sec:appendix_standard_cl}.

\section{Conclusion}

We introduced \method, a continual learning method that anchors representations in the sparse feature space of a pre-trained Sparse Autoencoder.
By decomposing dense activations into a sparse, decoupled feature space, \method enables more targeted preservation of previously learned knowledge while maintaining plasticity for new tasks.
Our approach addresses the key technical challenges of operating in SAE feature space through a sparse-aware distillation loss on pre-ReLU features with active-feature masking and an adaptive $\lambda$ mechanism that automatically balances plasticity and stability.
\method achieves strong results on TRACE and a 4-task classification benchmark, outperforming compared methods across three model families.
Our work connects mechanistic interpretability with continual learning: the sparse, disentangled representations developed for understanding neural networks also serve as an effective inductive bias for controlling them.
Future work could explore multi-layer SAE features, scaling to longer task sequences, and leveraging SAE feature interpretability for task-aware regularization.

\clearpage
\section*{Limitations}

\method requires a pre-trained SAE per base model, adding a one-time training cost that limits applicability to architectures where SAEs are unavailable or perform poorly.
The method stores float16 MLP activations for anchor samples (${\sim}$0.8~GB per task), growing linearly with task count; for very long task sequences, compression strategies may be needed.
TRACE evaluates on a single fixed task order, and while multi-model consistency suggests robustness, the Standard CL benchmark shows moderate variance across orderings (SAE-FD AA ranges from 87.90\% to 88.55\%).
Finally, published baselines use different experimental settings (e.g., GORP uses 1 epoch/task vs.\ the official multi-epoch schedule, and SEEKR uses full fine-tuning vs.\ our LoRA), so cross-method comparisons should be interpreted with these caveats in mind.

\bibliography{references}

\clearpage
\appendix

\section{TRACE Benchmark Details}
\label{sec:appendix_benchmark}

TRACE \cite{wang2023trace} is a comprehensive continual learning benchmark for LLMs consisting of 8 heterogeneous tasks trained in a fixed order.
The benchmark is designed to test continual learning across diverse task types, including classification, generation, code completion, reasoning, and multilingual tasks.
Each task provides 5{,}000 training samples and 500 test samples.
Table~\ref{tab:benchmark_details} summarizes the tasks, and Table~\ref{tab:benchmark_examples} provides representative prompt-output examples.

All tasks use the LLaMA-2 instruction format: \texttt{[INST] \{prompt\} [/INST]}, where the task-specific prompt is wrapped in instruction tags.
During evaluation, the model generates up to 256 tokens with temperature 0.1.

\begin{table*}[t!]
\centering
\small
\begin{tabular}{llll}
\toprule
\textbf{Order} & \textbf{Task} & \textbf{Domain} & \textbf{Metric} \\
\midrule
T1 & C-STANCE & Stance detection on COVID-related claims & Accuracy \\
T2 & FOMC & Financial policy sentiment from Fed minutes & Accuracy \\
T3 & MeetingBank & Summarization of meeting transcripts & ROUGE-L \\
T4 & Py150 & Python code completion & Similarity \\
T5 & ScienceQA & Science multiple-choice QA with reasoning & Accuracy \\
T6 & NumGLUE-cm & Commonsense arithmetic word problems & Accuracy \\
T7 & NumGLUE-ds & Numerical data science reasoning & Accuracy \\
T8 & 20Minuten & German news text simplification & ROUGE-L \\
\bottomrule
\end{tabular}
\caption{TRACE benchmark task overview. Tasks are trained in fixed order T1--T8.}
\label{tab:benchmark_details}
\end{table*}

\begin{table*}[t!]
\centering
\small
\setlength{\tabcolsep}{3pt}
\begin{tabular}{p{1.5cm}p{10cm}p{3cm}}
\toprule
\textbf{Task} & \textbf{Prompt (abbreviated)} & \textbf{Output} \\
\midrule
C-STANCE & Given the target ``Wearing Face Masks'' and the text ``Masks are essential for public health during the pandemic'', what is the stance? Choose from: favor, against, none. & favor \\
\midrule
FOMC & Classify the following sentence from FOMC minutes as hawkish, dovish, or neutral: ``The Committee decided to raise the target range for the federal funds rate by 25 basis points.'' & hawkish \\
\midrule
Meeting\-Bank & Summarize the key decisions from the following meeting transcript: ``The board reviewed the quarterly budget report ...'' & The board approved the revised budget ... \\
\midrule
Py150 & Complete the following Python code: \texttt{def fibonacci(n): if n <= 1:} & \texttt{return n return fibonacci(n-1) + fibonacci(n-2)} \\
\midrule
ScienceQA & Question: Which property do these two objects have in common? Select the best answer. (A) smooth (B) scratchy (C) blue & A \\
\midrule
NumGLUE-cm & Sam had 9 dimes in his bank. His dad gave him 7 more dimes. How many dimes does Sam have now? & 16 \\
\midrule
NumGLUE-ds & There are 7 dogwood trees currently in the park. Park workers will plant 3 more dogwood trees today. How many dogwood trees will the park have? & 10 \\
\midrule
20Minuten & Vereinfache den folgenden Text: ``Die Regierung hat beschlossen, die Massnahmen zur Bek\"ampfung der Inflation zu versch\"arfen ...'' & Die Regierung versch\"arft die Massnahmen gegen die Inflation ... \\
\bottomrule
\end{tabular}
\caption{Representative prompt-output examples for each TRACE task.}
\label{tab:benchmark_examples}
\end{table*}

\section{Evaluation Metrics}
\label{sec:appendix_metrics}

TRACE uses task-specific evaluation metrics reflecting the diversity of its 8 tasks, following the official evaluation protocol from \citet{wang2023trace}.
Classification tasks (C-STANCE, FOMC, NumGLUE-cm, NumGLUE-ds) use strict string match accuracy: the model's generated output must exactly match the reference answer.
ScienceQA uses first-character match accuracy, extracting the answer letter from the beginning of the generated text.
Code completion (Py150) uses the fuzzywuzzy string similarity ratio (0--100, normalized to 0--1), which measures the edit distance between the generated and reference code.
Summarization tasks (MeetingBank, 20Minuten) use ROUGE-L F1 score computed via py-rouge, measuring the longest common subsequence between the generated and reference summaries.
All evaluations use temperature 0.1 sampling with a maximum of 256 new tokens and a maximum prompt length of 512 tokens.

\section{Hyperparameters and Training Details}
\label{sec:appendix_settings}

Table~\ref{tab:hyperparams} lists the complete hyperparameter settings for all experiments.
LoRA is applied to the query, key, value, and output projection matrices of all attention layers.
The per-task epoch schedule follows the official TRACE configuration, with longer schedules assigned to more complex tasks.
For SAE-FD, 200 anchor samples are captured per task and stored in float16, totaling approximately 5.6~GB across all 7 anchor sets (tasks 1--7).
The adaptive $\lambda$ mechanism uses a target contribution ratio of 15\%, meaning the distillation loss is maintained at approximately 15\% of the total training loss.

\begin{table}[t]
\centering
\small
\setlength{\tabcolsep}{4pt}
\begin{tabular}{ll}
\toprule
\textbf{Parameter} & \textbf{Value} \\
\midrule
\multicolumn{2}{l}{\textit{LoRA}} \\
Rank / Alpha / Dropout & 8 / 32 / 0.1 \\
Target modules & q, k, v, o\_proj \\
\midrule
\multicolumn{2}{l}{\textit{Training}} \\
Learning rate & 1e-4 (const.\ + 10\% warmup) \\
Optimizer & AdamW (wd 0.0) \\
Batch (effective) & 4 $\times$ 8 accum = 32 \\
Per-task epochs & [5,3,7,5,3,5,5,7] \\
Max seq.\ length & 1024 \\
Train / eval samples & 5000 / 500 per task \\
\midrule
\multicolumn{2}{l}{\textit{SAE-FD}} \\
Anchors per task & 200 \\
Target $\rho$ / EMA $\alpha$ & 0.15 / 0.05 \\
$\lambda_{\min}$ / $\lambda_{\max}$ & 0.2 / 10.0 \\
\midrule
\multicolumn{2}{l}{\textit{SAE Training}} \\
Dimensions ($d \to D$) & 4096 $\to$ 32768 \\
Type & Gated SAE \\
Training loss & MSE + L1 ($\lambda_{\ell_1}=10^{-3}$) \\
Optimizer & AdamW (lr 3e-4, cosine decay) \\
Epochs / Batch size & 30 / 128 \\
Variance explained & $>$99\% \\
\midrule
\multicolumn{2}{l}{\textit{Evaluation}} \\
Max prompt / new tokens & 512 / 256 \\
Temperature / do\_sample & 0.1 / True \\
\bottomrule
\end{tabular}
\caption{Complete hyperparameter settings.}
\label{tab:hyperparams}
\end{table}

\section{Adaptive $\lambda$ Dynamics}
\label{sec:appendix_lambda}

Table~\ref{tab:lambda_dynamics} shows the adaptive $\lambda$ values at the start and end of each task's training for LLaMA-2-7B-Chat.
The mechanism exhibits intuitive behavior aligned with forgetting risk: $\lambda$ is initialized near its maximum when a new task begins (high task loss drives up the ideal $\lambda$) and decays as the model converges.
Notably, task-dependent decay rates emerge automatically: short tasks like FOMC (3 epochs) see rapid decay to the minimum ($10.0 \to 0.20$), while longer and more complex tasks like MeetingBank (7 epochs) maintain elevated $\lambda$ throughout ($10.0 \to 4.80$), reflecting sustained forgetting risk from extended training.

\begin{table}[t]
\centering
\small
\begin{tabular}{lcc}
\toprule
\textbf{Task} & $\lambda_{\text{start}}$ & $\lambda_{\text{end}}$ \\
\midrule
T2 (FOMC, 3 epochs) & 10.0 & 0.20 \\
T3 (MeetingBank, 7 epochs) & 10.0 & 4.80 \\
T4 (Py150, 5 epochs) & 10.0 & 0.97 \\
T5 (ScienceQA, 3 epochs) & 7.5 & 3.3 \\
T6 (NumGLUE-cm, 5 epochs) & 1.6 & 0.20 \\
T7 (NumGLUE-ds, 5 epochs) & 4.9 & 0.44 \\
\bottomrule
\end{tabular}
\caption{Adaptive $\lambda$ dynamics for LLaMA-2-7B-Chat on TRACE.}
\label{tab:lambda_dynamics}
\end{table}

\section{Per-Task BWT Breakdown}
\label{sec:appendix_bwt}

Table~\ref{tab:bwt_breakdown} shows the per-task BWT for \method on TRACE (LLaMA-2-7B-Chat).
Five of seven previously learned tasks exhibit $|\text{BWT}| < 2$\%, with positive transfer observed on FOMC (+0.8\%) and NumGLUE-cm (+2.4\%).
The largest degradation occurs on ScienceQA ($-$3.6\%), a multi-step reasoning task requiring chain-of-thought generation, suggesting that compositional reasoning features may be harder to preserve than simpler classification or generation features.

\begin{table}[t]
\centering
\small
\begin{tabular}{lccc}
\toprule
\textbf{Task} & \textbf{Peak} & \textbf{Final} & \textbf{BWT} \\
\midrule
C-STANCE & 48.8 & 48.0 & $-$0.8 \\
FOMC & 69.8 & 70.6 & +0.8 \\
MeetingBank & 61.2 & 60.9 & $-$0.3 \\
Py150 & 48.4 & 48.3 & $-$0.1 \\
ScienceQA & 82.4 & 78.8 & $-$3.6 \\
NumGLUE-cm & 38.3 & 40.7 & +2.4 \\
NumGLUE-ds & 53.5 & 51.7 & $-$1.8 \\
\midrule
\textbf{Average} & & & $\mathbf{-0.46}$ \\
\bottomrule
\end{tabular}
\caption{Per-task BWT for \method on TRACE (LLaMA-2-7B-Chat). Values in \%.}
\label{tab:bwt_breakdown}
\end{table}

\section{Full Accuracy Matrix}
\label{sec:appendix_matrix}

Table~\ref{tab:accuracy_matrix} presents the full accuracy matrix for \method on TRACE (LLaMA-2-7B-Chat).
Each row shows performance on all tasks seen so far after training on the corresponding task.
The matrix demonstrates remarkable stability: most previously learned tasks maintain their performance throughout the entire 8-task sequence, with only minor fluctuations.
For example, FOMC accuracy remains in the 69.8--72.4\% range from task 2 through task 8, and MeetingBank stays within 60.5--61.2\% from task 3 onward.

\begin{table*}[t!]
\centering
\small
\begin{tabular}{l|cccccccc}
\toprule
& C-ST & FOMC & MB & Py150 & SciQA & N-cm & N-ds & 20Min \\
\midrule
After T1 & 48.8 & --- & --- & --- & --- & --- & --- & --- \\
After T2 & 45.8 & 69.8 & --- & --- & --- & --- & --- & --- \\
After T3 & 48.2 & 71.0 & 61.2 & --- & --- & --- & --- & --- \\
After T4 & 48.0 & 71.8 & 60.6 & 48.4 & --- & --- & --- & --- \\
After T5 & 46.4 & 71.4 & 60.5 & 48.5 & 82.4 & --- & --- & --- \\
After T6 & 47.0 & 72.4 & 60.5 & 47.7 & 78.8 & 38.3 & --- & --- \\
After T7 & 46.4 & 70.6 & 60.8 & 48.4 & 77.2 & 34.6 & 53.5 & --- \\
After T8 & 48.0 & 70.6 & 60.9 & 48.3 & 78.8 & 40.7 & 51.7 & 22.5 \\
\bottomrule
\end{tabular}
\caption{Full accuracy matrix (\%) for \method on TRACE (LLaMA-2-7B-Chat). Each row shows performance on all tasks seen so far after training on that row's task.}
\label{tab:accuracy_matrix}
\end{table*}

\section{Published Baselines by Model}
\label{sec:appendix_baselines}

Table~\ref{tab:baselines_full} provides comprehensive comparisons with all published baselines for Mistral-7B and Vicuna-7B.
These baselines are reported from the original papers using their respective experimental configurations.
\method outperforms all published LoRA-based baselines on both models, and on Vicuna-7B achieves competitive accuracy with SEEKR-1\% while using parameter-efficient LoRA instead of full fine-tuning.

\begin{table}[t]
\centering
\small
\setlength{\tabcolsep}{3.5pt}
\begin{tabular}{llcc}
\toprule
\textbf{Model} & \textbf{Method} & \textbf{AA}$\uparrow$ & \textbf{BWT}$\uparrow$ \\
\midrule
\multirow{9}{*}{\shortstack[l]{Mistral\\-7B}}
& SeqLoRA & 46.94 & $-$11.41 \\
& L2P \cite{wang2022l2p} & 49.32 & $-$5.34 \\
& DualPrompt \cite{wang2022dualprompt} & 51.14 & $-$6.13 \\
& HiDe-PET \cite{wang2024hidepet} & 51.81 & $-$6.25 \\
& O-LoRA \cite{wang2023olora} & 52.02 & $-$8.13 \\
& GEM \cite{lopezpaz2017gem} & 52.32 & $-$6.01 \\
& EWC \cite{kirkpatrick2017ewc} & 52.45 & $-$5.98 \\
& TreeLoRA \cite{qian2025treelora} & 54.77 & $-$3.77 \\
& \textbf{\method (ours)} & \textbf{57.80} & $\mathbf{-1.63}$ \\
\midrule
\multirow{6}{*}{\shortstack[l]{Vicuna\\-7B}}
& LwF \cite{li2017lwf} & 41.19 & $-$5.54 \\
& EWC \cite{kirkpatrick2017ewc} & 41.88 & $-$15.57 \\
& O-LoRA \cite{wang2023olora} & 43.42 & $-$6.27 \\
& DER++ (1\%) \cite{buzzega2020der} & 49.01 & $-$9.04 \\
& SEEKR (1\%) \cite{he2024seekr} & 55.78 & $-$2.64 \\
& \textbf{\method (ours)} & \textbf{53.50} & $\mathbf{-0.69}$ \\
\bottomrule
\end{tabular}
\caption{Full TRACE baseline comparisons for Mistral-7B and Vicuna-7B. AA and BWT are reported in \%. \method achieves the best BWT on both models.}
\label{tab:baselines_full}
\end{table}

\clearpage
\section{Computational Overhead}
\label{sec:appendix_efficiency}

To quantify the computational overhead of \method, we compare per-task training time and performance against SeqLoRA and SeqLoRA+ER on Vicuna-7B (5{,}000 samples/task, identical epoch schedule, single NVIDIA H100 GPU).
Table~\ref{tab:efficiency} reports training time (relative to SeqLoRA), peak scores (immediately after training on each task), final retained scores (after all 8 tasks), and forgetting (Drop = Final $-$ Peak).

For T1 (C-STANCE), \method has zero overhead (1.00$\times$) because no anchor buffer exists yet.
From T2 onward, \method adds one forward pass on a sampled anchor batch per training step, resulting in consistent ${\sim}$2$\times$ overhead.
This overhead does not grow with the number of previous tasks, as \method samples a single fixed-size anchor batch regardless of buffer size.
The total overhead is 1.85$\times$ relative to SeqLoRA.
SeqLoRA+ER incurs additional overhead from replaying previous-task data, which grows as the replay buffer expands with each task.

All three methods achieve comparable peak performance on each task, confirming that neither feature distillation nor replay impedes new-task learning.
The critical difference is retention: SeqLoRA catastrophically forgets early tasks (MeetingBank: 62.6\% $\to$ 16.0\%), SeqLoRA+ER partially mitigates this (60.3\% $\to$ 56.6\%), while \method retains nearly all peak performance (64.2\% $\to$ 62.2\%).
\method achieves the best overall AA (53.4\%) and BWT ($-$0.7\%), outperforming SeqLoRA+ER (49.7\% AA, $-$3.8\% BWT) at 1.85$\times$ the cost of SeqLoRA.
Additionally, the frozen SAE adds a fixed 256~MB storage cost, and per-task anchor activations add ${\sim}$0.8~GB per task in float16.

\begin{table*}[t!]
\centering
\small
\setlength{\tabcolsep}{2.2pt}
\begin{tabular}{l ccc ccc ccc ccc}
\toprule
& \multicolumn{3}{c}{\textbf{Time ($\times$Seq)}} & \multicolumn{3}{c}{\textbf{Peak (\%)}} & \multicolumn{3}{c}{\textbf{Final (\%)}} & \multicolumn{3}{c}{\textbf{Drop}} \\
\cmidrule(lr){2-4} \cmidrule(lr){5-7} \cmidrule(lr){8-10} \cmidrule(lr){11-13}
\textbf{Task} & \textbf{Seq} & \textbf{+ER} & \textbf{FD} & \textbf{Seq} & \textbf{+ER} & \textbf{FD} & \textbf{Seq} & \textbf{+ER} & \textbf{FD} & \textbf{Seq} & \textbf{+ER} & \textbf{FD} \\
\midrule
T1 C-ST & 1.00 & 1.00 & 1.00 & 51.0 & 51.6 & 47.4 & 43.4 & 48.0 & 48.2 & \footnotesize{$-$7.6} & \footnotesize{$-$3.6} & \footnotesize{+0.8} \\
T2 FOMC & 1.00 & 1.03 & 2.07 & 70.8 & 70.0 & 71.4 & 49.6 & 67.3 & 71.6 & \footnotesize{$-$21.2} & \footnotesize{$-$2.6} & \footnotesize{+0.2} \\
T3 MB & 1.00 & 1.06 & 1.97 & 62.6 & 60.3 & 64.2 & 16.0 & 56.6 & 62.2 & \footnotesize{$-$46.6} & \footnotesize{$-$3.7} & \footnotesize{$-$2.0} \\
T4 Py150 & 1.00 & 1.10 & 2.04 & 46.6 & 45.1 & 45.3 & 36.5 & 42.7 & 45.4 & \footnotesize{$-$10.1} & \footnotesize{$-$2.4} & \footnotesize{+0.1} \\
T5 SciQA & 1.00 & 0.87 & 1.46 & 80.0 & 88.4 & 78.4 & 71.4 & 79.8 & 75.4 & \footnotesize{$-$8.6} & \footnotesize{$-$8.6} & \footnotesize{$-$3.0} \\
T6 N-cm & 1.00 & 1.23 & 2.08 & 44.4 & 34.6 & 46.9 & 33.3 & 32.1 & 46.9 & \footnotesize{$-$11.1} & \footnotesize{$-$2.5} & \footnotesize{+0.0} \\
T7 N-ds & 1.00 & 1.28 & 2.08 & 54.2 & 55.1 & 58.2 & 51.4 & 51.7 & 57.2 & \footnotesize{$-$2.8} & \footnotesize{$-$3.4} & \footnotesize{$-$1.0} \\
T8 20Min & 1.00 & 1.24 & 1.96 & 19.9 & 19.5 & 21.1 & 19.9 & 19.5 & 21.1 & \footnotesize{+0.0} & \footnotesize{+0.0} & \footnotesize{+0.0} \\
\midrule
\textbf{Overall} & \textbf{1.00} & \textbf{1.12} & \textbf{1.85} & \textbf{53.7} & \textbf{53.1} & \textbf{54.1} & \textbf{40.2} & \textbf{49.7} & \textbf{53.4} & \textbf{$-$15.4} & \textbf{$-$3.8} & \textbf{$-$0.7} \\
\bottomrule
\end{tabular}
\caption{Per-task training time, peak scores, final retained scores, and forgetting on TRACE (Vicuna-7B). Seq = SeqLoRA, +ER = SeqLoRA+ER, FD = \method.}
\label{tab:efficiency}
\end{table*}

\section{Standard CL Benchmark}
\label{sec:appendix_standard_cl}

Table~\ref{tab:standard_cl} presents results on a 4-task text classification benchmark, averaged over 3 random task orders.
\method achieves 88.25\% AA with $-$0.89\% BWT.
Our results exceed published baselines (SAPT-LoRA 81.1\%, GORP 78.6\%), though the gap between our SeqLoRA (85.6\%) and published baselines suggests differences in evaluation protocol, so cross-paper comparisons should be interpreted with caution.

\begin{table}[t]
\centering
\small
\setlength{\tabcolsep}{4pt}
\begin{tabular}{lcc}
\toprule
\textbf{Method} & \textbf{AA}$\uparrow$ & \textbf{BWT}$\uparrow$ \\
\midrule
\multicolumn{3}{l}{\textit{Published baselines (LLaMA-2-7B)}} \\
O-LoRA \cite{wang2023olora} & 76.1 & --- \\
N-LoRA \cite{yang2025nlora} & 77.6 & --- \\
GORP \cite{wang2025gorp} & 78.6 & --- \\
InfLoRA \cite{liang2024inflora} & 79.6 & --- \\
DATA \cite{liao2025data} & 79.8 & --- \\
SLAO \cite{qiao2025slao} & 80.4 & --- \\
SAPT-LoRA \cite{zhao2024sapt} & 81.1 & --- \\
\midrule
\multicolumn{3}{l}{\textit{Our results (LLaMA-2-7B, 3 orders)}} \\
SeqLoRA & 85.62 & $-$4.71 \\
\method & \textbf{88.25} & $\mathbf{-0.89}$ \\
\bottomrule
\end{tabular}
\caption{Standard CL benchmark results (4 tasks, LLaMA-2-7B-Chat). AA and BWT are reported in \%.}
\label{tab:standard_cl}
\end{table}

\section{SAE Reconstruction Quality Under Continual Fine-Tuning}
\label{sec:appendix_sae_quality}

A key concern with \method is whether the frozen SAE remains a faithful decomposition of model activations as LoRA fine-tuning progressively shifts the representations across tasks.
If the SAE's reconstruction quality degrades substantially, the feature distillation signal may become unreliable.
To investigate this, we evaluate the frozen Vicuna-7B SAE's reconstruction quality on model activations collected after each of the 8 TRACE tasks, using 500 random samples per evaluation.

\begin{figure}[t]
\centering
\includegraphics[width=\columnwidth]{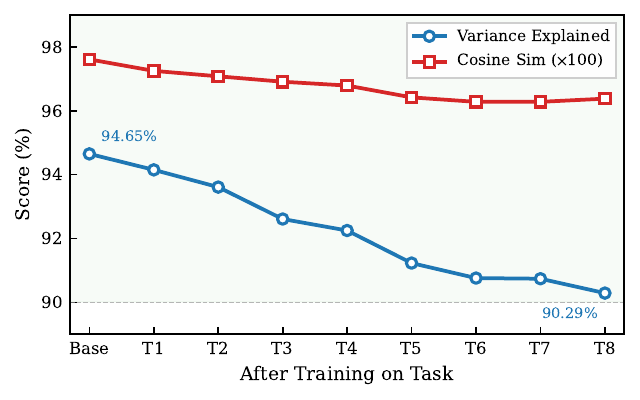}
\caption{Frozen SAE reconstruction quality on Vicuna-7B activations after each TRACE task. Variance explained decreases gradually from 94.65\% to 90.29\% ($-$4.36\%) over 8 tasks. Cosine similarity (scaled $\times$100) remains above 96\% throughout, confirming that the frozen SAE provides a reliable feature decomposition under continual LoRA fine-tuning.}
\label{fig:sae_quality}
\end{figure}

Figure~\ref{fig:sae_quality} tracks two key metrics across the 8-task sequence: variance explained (proportion of activation variance captured by the SAE reconstruction) and cosine similarity between original and reconstructed activations.
Note that the SAE achieves $>$99\% variance explained on its own held-out training data (diverse general text); the lower starting point of 94.65\% on the base model reflects a domain gap between the SAE's training distribution and the TRACE benchmark data, which includes specialized domains such as Python code, financial text, and German text that are underrepresented in the SAE training corpus.
From this baseline, variance explained decreases gradually to 90.29\% after task 8, with only 4.36 percentage points of additional degradation attributable to LoRA fine-tuning across the entire 8-task sequence.
Cosine similarity remains above 0.96 throughout, indicating that the SAE feature space remains structurally intact.

These results justify the use of a frozen SAE for feature distillation: LoRA fine-tuning introduces modest representational shifts that reduce reconstruction fidelity only marginally.
The SAE continues to provide a meaningful sparse decomposition of model activations even after 8 sequential tasks, supporting the reliability of the distillation signal throughout the continual learning process.

\end{document}